\documentclass[a4paper]{article}

\usepackage{INTERSPEECH2022}
\usepackage{amsmath,graphicx}
\usepackage{xcolor}
\usepackage{amsfonts}
\usepackage{amssymb,amsmath,bm}
\usepackage{multirow}
\usepackage{booktabs}
\usepackage{multicol}
\usepackage{array}
\usepackage{enumitem}
\usepackage{multirow, makecell}
\usepackage[mathscr]{euscript}
\usepackage{float}
\usepackage{hyperref}
\usepackage{tablefootnote}
\usepackage{graphics}
\usepackage{cite}

\usepackage{todonotes} %

\newcommand{\Zhehuai}[2][]{}
\newcommand{\andrew}[2][]{}
\newcommand{\Andrew}[2][]{}
\newcommand{\ankur}[2][]{}
\newcommand{\Ankur}[2][]{}
\newcommand{\ngyuzh}[2][]{}
\newcommand{\Ngyuzh}[2][]{}
\newcommand{\bhuv}[2][]{}
\newcommand{\method}{Maestro}

\newcommand{\myparagraph}[1]{\vspace{.4em} \noindent \textbf{#1}\ }

\global\hyphenpenalty=5000

\ninept

\title{
MAESTRO: Matched Speech Text Representations through Modality Matching
}
\name{Zhehuai~Chen,~Yu~Zhang,~Andrew~Rosenberg,~Bhuvana~Ramabhadran, \\ 
Pedro~Moreno,~Ankur~Bapna,~Heiga~Zen}
\address{Google, Inc.}
\email{\{zhehuai,ngyuzh,rosenberg,bhuv,pedro,ankurbpn,heigazen\}@google.com}

\begin{document}

\maketitle
\begin{abstract}

We present \method{}, a self-supervised training method to unify representations learnt from speech and text modalities. 
Self-supervised learning from speech signals aims to learn the latent structure inherent in the signal, while  self-supervised learning from text attempts to capture lexical information. Learning aligned representations from unpaired speech and text sequences is a challenging task.  Previous work either implicitly enforced the representations learnt from these two modalities to be aligned in the latent space through multi-tasking and parameter sharing or explicitly through conversion of modalities via speech synthesis. While the former suffers from interference between the two modalities, the latter introduces additional complexity. In this paper, we propose \method{}, a novel algorithm to learn unified representations from both these modalities simultaneously that can transfer to diverse downstream tasks such as Automated Speech Recognition (ASR) and Speech Translation (ST). \method{} learns unified representations through sequence alignment, duration prediction and  matching embeddings in the learned space through an aligned masked-language model loss.  We establish  a new state-of-the-art (SOTA)
on VoxPopuli multilingual ASR with a \textbf{8\%} relative reduction in Word Error Rate (WER), multi-domain SpeechStew ASR (\textbf{3.7\%} relative) and 21 languages to English multilingual ST on CoVoST 2 with an improvement of \textbf{2.8} BLEU averaged over 21 languages.

\end{abstract}

\section{Introduction}
In recent years, self-supervised learning (aka pretraining) has become an effective and scalable paradigm to learn representations from speech and text. These representations of language learnt without labeled data have shown impressive performance in multiple speech and natural language processing tasks ~\cite{devlin2018bert,xue2020mt5,baevski2020wav2vec,conneau2019unsupervised}. Auto-regressive, contrastive and multi-task learning objectives have been proposed in the literature to pre-train such models~\cite{ao2021speecht5, chung2020vector, baevski2020wav2vec, zhang2020pushing, hsu2021hubert2, chung2021w2v}, followed by fine-tuning with labeled data to the task of interest, namely, speech recognition, translation, voice conversion and spoken language understanding. These diverse tasks require that the learned representations capture acoustic, prosodic, speaker and linguistic characteristics as well as the semantics of the spoken language. Thus, joint pre-training of representations from both speech and text modalities is a natural extension for improved generalization.

Central to the problem of jointly learning shared speech and text representation is the challenging task of combining two very different modalities. Recent methods proposed in literature have addressed this challenge through two very different approaches. One approach uses multi-task training which trains a single model with different objectives for each modality~\cite{bapna2021slam,ao2021speecht5,bapna2022mslam}, while another uses modality conversion where text is directly converted to speech~\cite{chen2021injecting} via text-to-speech (TTS) approaches or viceversa~\cite{kuhn1990cache} prior to self-supervised training. 

\method{} (Section ~\ref{sec:approach}) lies at the intersection of these two lines of work. It retains the advantage of a common latent space and alignment between the speech and text representations without explicitly converting text to speech or viceversa, thereby learning a computationally efficient and aligned representation of the two modalities. 
The main contributions of this paper are:
\begin{itemize}[leftmargin=1em]
\item A novel modality matching %
algorithm, \method{}, that can effectively use small amounts of transcribed speech data to unify representations learnt from unlabeled speech and text.
\item A new algorithm to use additional sources of supervision, namely, Machine Translation (MT) and Speech Translation (ST), to improve multilingual joint representations learnt during pre-training.
\item Establishing a new state-of-the-art (SOTA) on the VoxPopuli multilingual ASR task with a \textbf{8\%} relative reduction in WER and on CoVoST-2 21 languages-to-English speech translation (ST) with an absolute improvement of \textbf{2.8} BLEU.

\end{itemize}

\section{Related Work}
With the convergence of architectures ~\cite{vaswani2017attention,gulati2020conformer,jaegle2021perceiver} and objectives~\cite{chung2021w2v,baevski2022data2vec} used for learning representations in speech and natural language processing, there has been growing interest in learning joint speech and text representations.
One set of approaches pursue multi-task training~\cite{bapna2021slam,ao2021speecht5}. However, given the very different natures of the two modalities and their training objectives, these approaches can suffer from interference and capacity limitations.
In contrast, alternative approaches utilize Text-To-Speech (TTS) synthesis to augment natural speech data, avoiding inter-modal interference and improving downstream performance, at the cost of requiring a complete TTS model~\cite{chen2021injecting,tts4pretrain2} despite utilizing only the intermediate feature representations between text and speech for augmentation~\cite{renduchintala2018multi,wang2020improving,chen2021injecting,uenodata}. \method{} bridges these approaches by implicitly aligning text representations with speech in the latent representation space. Utilizing recent advances in end-to-end speech synthesis that allow explicit duration modeling and control~\cite{elias2021parallel,ren2019fastspeech}, we show that we can use the duration predictions for alignment while enforcing similarity to natural speech representations.  %

Representation learning approaches can be extended to multilingual settings for text~\cite{conneau2019unsupervised,xue2020mt5}, speech~\cite{babu2021xls} and joint models~\cite{bapna2022mslam}. Learnt representations can also be improved by utilizing additional supervised data, joint unsupervised and supervised training on transcribed speech~\cite{bai2022just} or paired Masked Language Modeling (MLM) objectives on Machine Translation (MT) ~\cite{lample2019cross} or Speech Translation (ST) data~\cite{zheng2021fused}. We apply \method{} to train both, monolingual and massively multilingual models of speech and text. We also develop approaches to utilize cross-lingual supervision from MT and ST data during pre-training. Our work improves upon existing work on utilizing TTS models for speech representation learning by: (i) Developing an iterative and self-supervised approach for duration modeling, (ii) Circumventing the need for explicit speaker and prosody modeling by aligning resampled text representations with speech, (iii) Extending joint speech and text representation learning to massively multilingual datasets and tasks.

\section{Proposed method}
\label{sec:approach}

\subsection{Architecture}
\begin{figure}[hbt!]
  \centering
\includegraphics[width=1.0\linewidth]{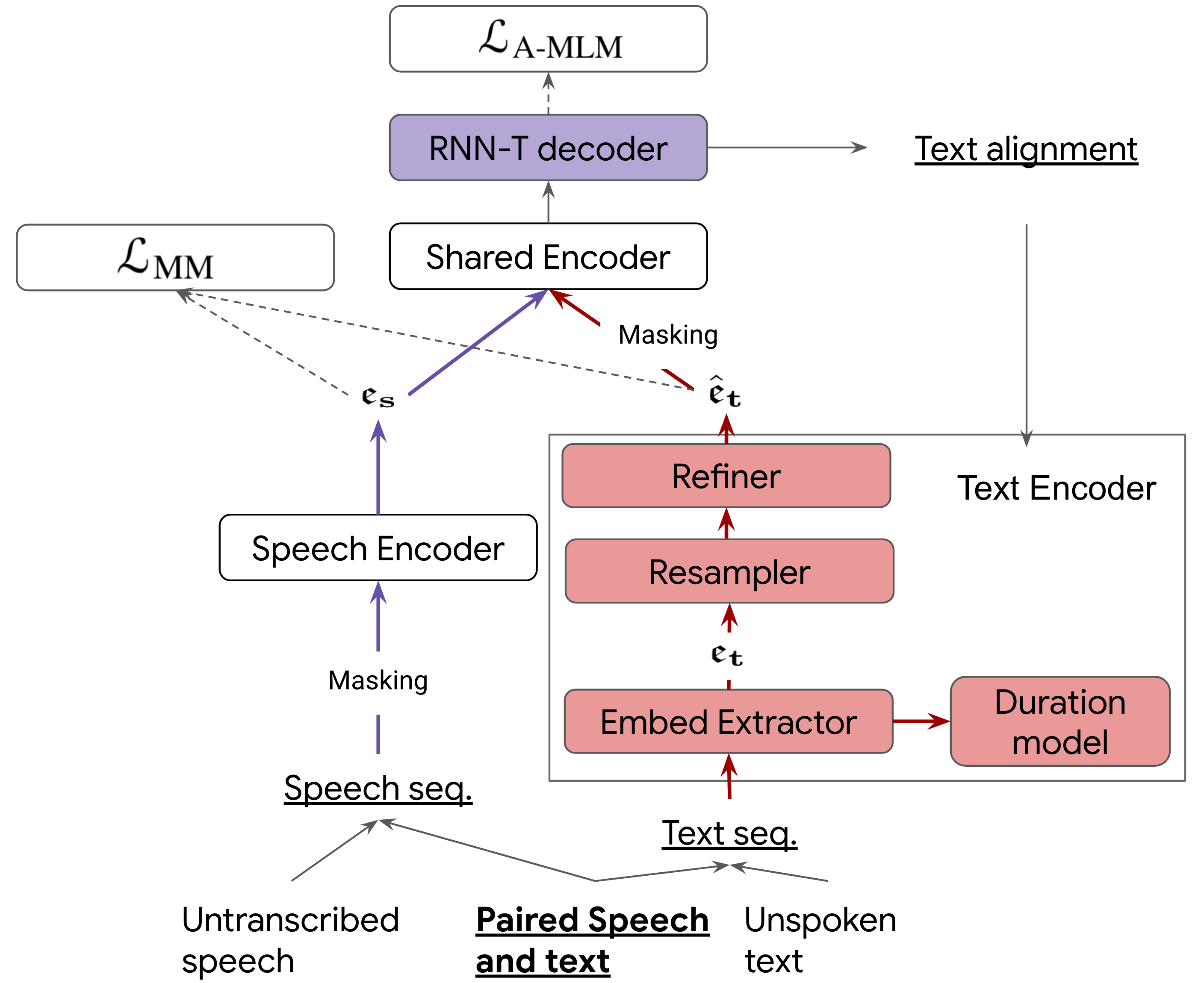}
    \caption{Proposed architecture of \method{} to learn unified representations from speech and text. The purple and red boxes denote differences from mSLAM~\cite{bapna2022mslam}. The {\it Text Encoder} block utilizes alignments to explicitly resample text representations $\mathfrak{e}_\mathbf{t}$ to match speech encoder output $\mathfrak{e}_\mathbf{s}$.
    }
    \label{fig:framework}
    \vspace{-3mm}
\end{figure}

The proposed framework to pretrain one model from untranscribed speech, unspoken text and any available labeled data (paired speech and text) is presented in Figure~\ref{fig:framework}. It  comprises of separate speech and text encoders to encode speech and text input signals, and a shared, multi-modal latent space encoder to learn the joint representation from these two modalities. The modality matching objective, $\mathscr{L}_\text{MM}$ is used to explicitly enforce speech-text modality matching 
while the $\mathscr{L}_\text{A-MLM}$ objective in the joint embedding space is used to learn representations from unspoken text. %
The framework allows for task-dependent additional RNN-T (or any other) decoders that predict grapheme, phoneme, word-piece or word targets from the shared representations.

\subsection{Modality Matched Training}
\label{sec:moma}

In this Section, we describe the 
overall training process using available paired speech and text data, unspoken text and modality matching. We propose a {\em modality matching mechanism} (MM) to explicitly unify the representation space learned from text and speech. Specifically, we try to link the two modalities in the phonetic level and learn a mapping between text units and  speech representations independent of speaker, prosody and other acoustic characteristics. 

\subsubsection{Learning Initial Embeddings}
The speech encoder $\theta_s$ is trained on speech input using a within-utterance contrastive loss focusing on
local acoustic and phonetic  information~\cite{baevski2020wav2vec,chung2021w2v}.
In contrast, the text embedding extractor $\theta_t$ captures the embedded lexical information from the input text signal.

\subsubsection{Learning Aligned Embeddings}
The initial speech  $\mathfrak{e}_\mathbf{s}$ and text $\mathfrak{e}_\mathbf{t}$ embeddings learned are not expected to be aligned. Merging the two independently learned representations without attention to alignment can result in poor knowledge sharing and require extra model capacity.  We hypothesize that this also limits the ability to jointly learn from the two tightly-coupled modalities representing any language. 
The self-alignment process described below learns alignments from the model itself in an iterative fashion. 

\textbf {Paired Speech and Text input}: We utilize an RNN-T decoder (See Figure~\ref{fig:framework}) and the available labeled data to first learn an RNN-T model.%
When learning from paired speech and text, the {\it Text Encoder} block in the figure uses this RNN-T model to generate alignments between the predicted text targets $\mathbf{t}$ and the speech encoder output $\mathfrak{e}_\mathbf{s}$. The {\it Resampler} and {\it Refiner} layers replicate the initially learned text embeddings to match the duration of the speech embedding using this alignment information and a Mean-Squared Error (MSE) training objective given in Equation~\ref{eqn:mse} below. 
\begin{eqnarray}
\label{eqn:mse}
\begin{aligned}
&\mathfrak{e}_\mathbf{s} = \theta_{s}(\mathbf{s}),\ \mathfrak{e}_\mathbf{t} = \theta_{t}(\mathbf{t}), \quad
(\mathbf{t},\mathbf{s})  \in \mathcal{X}_\text{paired}\\
& \hat{\mathfrak{e}}_\mathbf{t}=\theta_\text{Refiner}\Big(\text{Resample}\big(\mathfrak{e}_\mathbf{t}, \text{Align}_{\tt Rnnt}(\mathfrak{e}_\mathbf{s}, \mathbf{t})\big)\Big) \\
& \mathscr{L}_\text{MM} = \text{MSE}(\mathfrak{e}_\mathbf{s}, \hat{\mathfrak{e}}_\mathbf{t}) + \mathscr{L}_{\tt Rnnt} (\mathbf{t} \mid \mathfrak{e}_\mathbf{s}) 
\end{aligned}
\end{eqnarray}

\textbf {Unpaired text input}:
When learning from unspoken text, the speech-text alignment information is unavailable. We substitute it with durations predicted from a duration prediction model in a fashion similar to speech synthesis~\cite{elias2021parallel}. This model is  trained on any available paired data to predict the duration of each token. The predicted duration on unspoken text is subsequently used to resample the  initially learned text embeddings.

\subsubsection{Aligned Masked Language Model training objective}
With the introduction of speech-aligned and resampled text embeddings ($\hat{\mathfrak{e}}_\mathbf{t}$), we can use the same RNN-T loss for both, unpaired text (aligned embedding $\hat{\mathfrak{e}}_\mathbf{t}$, text) and paired (speech embedding $\mathfrak{e}_\mathbf{s}$, text) data. We replace the original MLM/BERT loss used in prior work~\cite{devlin2018bert,bapna2022mslam} with the {\em aligned masked language model} training objective ($\mathscr{L}_\text{A-MLM}$). This is the RNN-T loss applied over the masked, resampled text embeddings with masking in frequency and time domain similar to SpecAugment~\cite{park2019specaugment}. This new objective allows for the use of the same RNN-T objective on  speech embedding or unspoken text with no associated speech embedding. Motivated by~\cite{tts4pretrain2} to enforce consistency between text and speech modalities, we also include the A-MLM loss when training with paired speech and text input. We summarize this training in Equation~\ref{eqn:amlm}.

\begin{align}
&\mathfrak{e}_\mathbf{t} = \theta_{t}(\mathbf{t}), \hat{\mathfrak{e}}_\mathbf{t}=\theta_\text{Refiner}\Big(\text{Resample}\big(\mathfrak{e}_\mathbf{t}, \theta_{\tt Duration}(\mathfrak{e}_\mathbf{t})\big)\Big) \nonumber\\
& \mathscr{L}_\text{A-MLM} = \mathscr{L}_{\tt Rnnt} \Big(\mathbf{t} \mid \text{Mask}(\hat{\mathfrak{e}}_\mathbf{t})\Big)
\label{eqn:amlm}, \quad \mathbf{t}  \in \mathcal{X}_\text{text}
\end{align}

\subsection{Untranscribed Speech}
Similar to W2v-BERT~\cite{chung2021w2v}, we use contrastive loss on the speech encoder outputs and a masked language model (MLM) loss on the shared encoder output to pretrain our model on untranscribed speech data.

\section{Data and Experimental Setup}
\subsection{Pre-training Data}
\label{sec:pretrain}
We explore pre-training \method{} on both monolingual (English) and multilinugal scenarios as well as ASR and translation tasks.

\myparagraph{Monolingual ASR:} We use unlabeled English-only speech from the LibriLight corpus~\cite{kahn2020libri} to pre-train monolingual models. Our unspoken text corpus consists of the text corpora from TEDLIUM~\cite{rousseau2012ted} and Librispeech~\cite{panayotov2015Librispeech}. We use SpeechStew~\cite{chan2021speechstew} as the labeled, paired speech and text corpus. %

\myparagraph{Multilingual ASR:} Following~\cite{bapna2022mslam}, we use 429k hours of public unlabeled speech corpora: VoxPopuli~\cite{wang2021voxpopuli}, CommonVoice~\cite{ardila2019common}, MLS~\cite{pratap2020mls} and BABEL~\cite{gales2014speech} to pre-train  multilingual ASR models. Our transcribed data draws from VoxPopuli transcribed dataset (VP-P), MLS and Babel, following~\cite{bapna2022mslam}. We use three different setups to study the effectiveness of unlabeled text corpora: (1) VoxPopuli text dataset (VP-T, 3GB) %
(2) mC4~\cite{xue2020mt5} spanning 101 languages (15TB)
and (3) VP-T and mC4.

\myparagraph{Speech Translation:} We explore the use of MT and ST supervised data in pretraining for downstream tasks such as ST (Section~\ref{sec:setup-st}). 
We use the source speech and text data from CoVoST 21 languages to English ST corpus as transcribed speech. We concatenate paired MT text sequences from CoVoST, WMT and TED datasets\cite{bapna2022mslam} and use them as unspoken text for pretraining.

\subsection{Architecture Details}
\myparagraph{Speech encoder and shared encoder}: The encoders in Figure~\ref{fig:framework} are a stack of ``Conformer blocks''.  We use the Conformer XL architecture described in~\cite{chen2021injecting} with $24$ layers of full-context Conformer blocks (600M parameters) where 6 of the lower layers are assigned as speech encoder and the upper 18 layers are the speech-text shared encoder. 

\myparagraph{Text encoder}: The text embedding extractor $\theta_t$(Section~\ref{sec:moma}) includes 3 convolutional layers of $512$ filters with kernel size $(5, 1)$, followed by a 6-layer Transformer with positional embedding.  
The upsampling is done by copying the original text embedding to the target length of  specified duration with positional embeddings to capture frame positions within text units as described in ~\cite{elias2021parallel}. %
The Refiner $\theta_\text{Refiner}$ includes 2 layers of 8-headed self-attention blocks with $17\times 1$ lightweight convolutions~\cite{wu2019pay}. The duration model includes four blocks of $3\times 1$ lightweight convolutions taking the original text embedding to predict the duration.  

\myparagraph{RNN-T decoder}: We use a 2-layer, 1280-dim LSTM with a joint network of 640 dims
as the RNN-T decoder. While we use an RNN-T decoder in this paper, the proposed framework allows for the use of any decoder (CTC, LAS, etc.).
We use both phoneme and  grapheme decoders for monolingual pre-training. %
When switching to multilingual pre-training,
we use a single decoder with 4k sentence-pieces as targets to cover 101 languages in the mC4 dataset.  We also consider variants using grapheme and phoneme targets with vocabulary sizes of 100 and 256 respectively on the VoxPopuli corpora for comparison.

\subsection{Pretraining hyper-parameters}
We include untranscribed speech, unspoken text, transcribed speech in each batch with a fixed ratio. 
To stabilize training, we use, (1) exponential-moving-averaged (EMA) with decay rate $0.9999$ to all prediction steps during alignment of transcribed speech, duration prediction and resampling of unspoken text; %
(2) a curriculum learning schedule to start from untranscribed speech-only training,  include transcribed speech after 500k steps and unspoken text after another 15k steps. The joint training of three types of data lasts for another 300K steps  with a learning rate schedule and optimizer given in~\cite{zhang2020pushing}.  We use an effective batch size of (256, 512, 256)  for (untranscribed speech, unspoken text and paired data) in SpeechStew. It increases to (1024, 1024, 256) in VoxPopuli and (1024, 8192, 512) when including mC4 to increase text throughput.  
\bhuv{Do we really need this?}
\ngyuzh{The batch size for unspoken text and pair data may still important.}

\subsection{Downstream Tasks}

\myparagraph{Speech Recognition} We evaluate our monolingual \method{} on multi-domain ASR using the SpeechStew~\cite{chan2021speechstew} task. All fine-tuning data and parameters follow ~\cite{zhang2020pushing}. We also initialize the RNNT decoder from Maestro, which means both encoder and decoder are pretrained.
Multilingual versions of \method{} are fine-tuned and evaluated on the 14-language VoxPopuli~\cite{wang2021voxpopuli} multilingual ASR task using grapheme targets~\cite{bapna2022mslam}.%

\myparagraph{Speech Translation}
\label{sec:setup-st}
Our multilingual speech translation evaluation on the CoVoST-2 multilingual XX-to-English
task that covers translation from 21 source languages into English~\cite{wang2020covost} closely follows~\cite{bapna2022mslam}. For joint fine-tuning we utilize the text translation datasets composed of CoVoST 2, WMT and TED.

\begin{figure*}[hbt!]
  \centering
    \caption{Breakdown of the improvement from \method{} on VoxPopuli and the comparison between phoneme and grapheme pretrain. The languages are sorted by the amount of data from high to low. 
    }
    \label{fig:breakdown}
\includegraphics[width=0.85\linewidth]{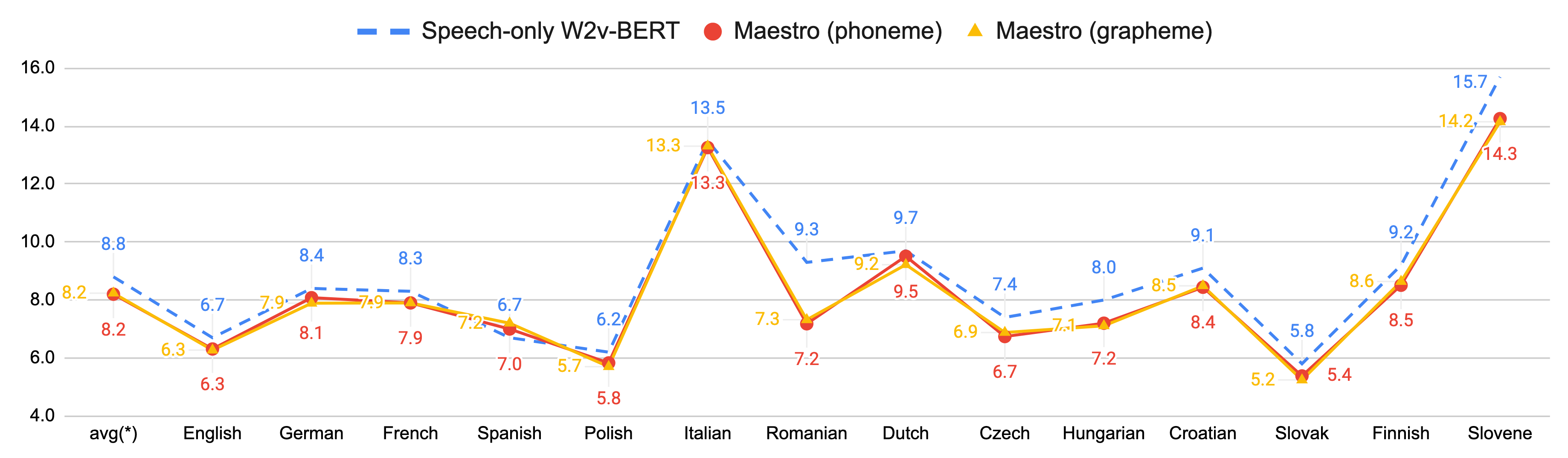}
    \vspace{-3mm}
\end{figure*}

\section{Experiments and Results}

\subsection{Monolingual Multi-domain Speech Recognition}

\begin{table}[tb]
\centering
 \caption{\label{tab:exp-mono} {Multi-domain ASR: SpeechStew results on 5 domains. All the models use 0.6B parameters. } }
 \resizebox{\columnwidth}{!}{
\begin{tabular}{l|cccccccc}
  \toprule
Method & \multicolumn{2}{c}{LS-test} & \multicolumn{2}{c}{AMI} &TED& \multicolumn{2}{c}{SWB} & CV  \\
   &\multicolumn{2}{r}{ {\em\scriptsize clean\ \ other}}  &\multicolumn{2}{c}{ {\em\scriptsize ihm\ \ \ \ \ \ sdm}}  &  &\multicolumn{2}{c}{ {\em\scriptsize swb\ \ \ \ \ \ chm}}&  \\
  \midrule
\footnotesize{Wav2vec2}        & 1.7                         & 3.3                              & 9.6                         & 23.8                        & 5.7                     & 4.9                         & 10.8                       & 8.5                    \\
\footnotesize{W2v-BERT}        & 1.6                         & 3                                & 9.1                         & 23.1                        & 5.4                     & 4.5                         & 9                          & 8.6                    \\
\footnotesize{\ \ + LM}            & \textbf{1.5}                         & 2.8                         \\
\midrule
\footnotesize{SLAM}          & 1.6                         & 3.1                              & 9.3                         & 23.5                        & 5.6                     & 4.6                         & 9.1                        & 8.6                    \\
\scriptsize{TTS4Pretrain2} & 1.6                         & 2.8                              & 8.7                         & \textbf{21.9}                        & 5                       & 4.5                         & 8.5                        & 8.4                    \\
  \midrule
\footnotesize{\textbf{\method{}}}         & \textbf{1.5}                         & 2.8                              & \textbf{8.5}                         & \textbf{21.9}                        & \textbf{4.9}                     & \textbf{4.3}                         & \textbf{8}                          & \textbf{8.1}                   \\
\footnotesize{\ \ + LM}            & \textbf{1.5}                         & \textbf{2.7}                      \\
\bottomrule
\end{tabular}
}
\vspace{-2em}
\end{table}

Table~\ref{tab:exp-mono} presents the results from fine-tuning monolingual \method{} on multi-domain ASR. The first two blocks present baselines from  speech-only pretraining methods, Wav2vec2 and W2v-BERT and previous speech-text pretraining methods, TTS4Pretrain 2.0 and SLAM~\cite{bapna2021slam} respectively. %
\method{} clearly outperforms speech-only pretraining and further improves over previous speech-text pretraining, TTS4Pretrain 2.0 (upto 6\% relative) and SLAM (upto 12\% relative) reduction in WER. 
Comparing \method{} with TTS4Pretrain 2.0, the text encoder here is learnt from all the SpeechStew data while TTS model used in TTS4Pretrain 2.0 is only trained on Librispeech. We believe this is the reason we see better performance.
On Librispeech, \method{} closes the gap with LM shallow fusion
(Row 3)  and yields additional wins with LM fusion (last row).
Both use a neural Conformer language model trained from Librispeech text corpora.

\subsection{Multilingual Speech Recognition}
\label{ss:masr}

\begin{table}[ht!]
 \caption{\label{tab:exp-voxp} {Multilingual ASR: VoxPopuli results on 14 languages.
  All  systems are finetuned with the same 1.3k hours of multilingual   VoxPopuli data (VP-P).
    In pretrain, 2.4k hours refer to VP-P+MLS+BABEL. \method{} uses 4k SPM during pretrain. (XLS-R uses 8k extra hours of untranscribed speech from VoxLingua~\cite{valk2021voxlingua107}.)       } }
    \vspace{-0.5em}
  \resizebox{\columnwidth}{!}{
\begin{tabular}{l|l|lll|c}
  \toprule
     & Model  & Paired              &      Speech           &    Text      & Avg  \\
Method   &  size &  (hours)             &      (hours)           &           &  WER \\
  \midrule
XLS-R~\cite{babu2021xls}     & 1B        & -  & 437k                    & -                       & 10.6                                    \\
W2v-BERT~\cite{chung2021w2v}  & 0.6B      & -   &   429k                           & -                       & 8.8                                   \\
  \midrule
mSLAM~\cite{bapna2022mslam}     & 0.6B      &2.4k  & 429k                   & mC4& 9.2                                     \\
mSLAM~\cite{bapna2022mslam}     & 2B       & 2.4k   & 429k                 & mC4  & 9.1                                     \\
  \midrule
\method{} & 0.6B    & 1.3k    & 429k                   & VP-T        & 8.2                                     \\
\method{} & 0.6B      & 2.4k   & 429k                & mC4  & 8.3                                     \\
\textbf{\method{}} & 0.6B      & 2.4k  & 429k     & \ \ +VP-T & \textbf{8.1}                                     \\
\bottomrule
\end{tabular}
}
\end{table}

We evaluate \method{} on multilingual speech recognition on the VoxPopuli benchmark in 
Table~\ref{tab:exp-voxp} and include several best-performing systems as comparison.
Using in-domain unspoken text (VP-T) and paired data (VP-P) provided by VoxPopuli release, the proposed method \method{} in the first row of the third block outperforms all the previous systems using either in-domain or out-domain unsupervised data. 
A slight degradation (Row 6) is seen with the inclusion of massive amounts of out-of-domain unspoken text such as mc4~\cite{xue2020mt5}.
Combining both mC4 and VP-T corpora (last row) yields the best performance, improving over speech-only W2v-BERT by 8\% relative and create a new  state of the art. 

To understand the improvement w.r.t. data sizes, Figure~\ref{fig:breakdown} illustrates the language-wise breakdown corresponding to Row 5 in Table~\ref{tab:exp-voxp}.
Most of languages show improvements including {\em Slovenian} with only 10 hours of supervised speech. This shows that the proposed method can learn meaningful speech-text representations even from small amounts of supervised data. Figure~\ref{fig:breakdown} also compares the use of phoneme and grapheme targets during pretraining. Notably, regardless of different units used in pretraining, graphemes are always used in the supervised finetune. The overall performances are similar. %

\subsection{Speech Translation}

We evaluate the potential of \method{} to serve as a foundation model for several speech recognition and understanding tasks. We use the same pretrained ASR encoder from Section~\ref{ss:masr} to evaluate speech translation performance.
Table~\ref{tab:exp-ast} demonstrates that \method{} also advances SOTA results on the speech translation CoVoST 2 benchmark.
With 0.6B parameters and the same pretrain and finetune data setups as mSLAM, \method{} (second last row) outperforms most of the previous systems except the larger, 2B parameter mSLAM model. 
To  minimize the mismatch in the text encoder outputs between pretraining and finetuning, we include MT and ST data into pretraining with the procedure described in Section~\ref{sec:pretrain} yielding the best performance (last row).
Overall, the resulting system outperforms mSLAM (0.6B) by  2.8 BLEU scores and creates a new SOTA on this benchmark with 30\% of the parameters.  We hypothesize that the extra efficiency stems from improved knowledge sharing between the two modalities that can be attributed to the proposed modality matching.

\begin{table}[tb!] 
\caption{\label{tab:exp-ast} {Speech Translation (ST):  CoVoST 2 X$\rightarrow$En results  on  21 language pairs. } }
 \vspace{-0.5em}
 \resizebox{\columnwidth}{!}
 {
\begin{tabular}{l|l|lcccc|c}
  \toprule
   & &  \multicolumn{5}{c|}{ Pretrain}              & \multicolumn{1}{l}{Avg } \\
    Method      &   Size         & \multicolumn{2}{l}{ Speech    \ Text}            & \multicolumn{3}{l|}{ASR\ \ ST\ \ MT}                        &   \multicolumn{1}{l}{BLEU}                     \\
  \midrule
   \multicolumn{7}{l}{{\em Finetune: ST-only; mBART decoder init~\cite{tang2021multilingual} }  }\\
  \midrule
XLS-R~\cite{babu2021xls}     & 1B         &  437k & $\times$                       &  $\times$       &  $\times$          & $\times$   & 19.3                                   \\
XLS-R~\cite{babu2021xls}     & 2B         &  437k & $\times$                      & $\times$     &  $\times$                      & $\times$    & 22.1                                     \\
  \midrule
   \multicolumn{4}{l}{{\em Finetune: ST and MT jointly  }}\\
   \midrule
   W2v-BERT\cite{chung2021w2v}  & 0.6B       &  429k &   $\times$                       &  $\times$       &  $\times$          & $\times$    & 21.0      \\
mSLAM~\cite{bapna2022mslam}     & 0.6B       & 429k    & mC4              & $\checkmark$    &  $\times$    & $\times$          & 22.4                                     \\
mSLAM~\cite{bapna2022mslam}     & 2B         &  429k    & mC4              & $\checkmark$    &  $\times$   & $\times$           & 24.8                                     \\
  \midrule
\method{} & 0.6B       &  429k    & mC4              & $\checkmark$    & $\times$    &  $\times$       & 24.3                                     \\
\textbf{\method{}} & 0.6B       & 429k   & mC4 & $\checkmark$ & $\checkmark$ & $\checkmark$    & \textbf{25.2}      \\                              
\bottomrule
\end{tabular}
}
\vspace{-2em}
\end{table}

\section{Conclusions}

We have described \method{}, a new technique for joint speech and text representation learning that outperforms the state of the art on speech recognition and speech translation tasks.  \method{} addresses the joint representation problem by first aligning text to speech and then training a text representation to match a W2v-BERT speech representation. This results in significant  improvements  by 8\% on ASR and 2.8 BLEU  on ST.

\bibliographystyle{IEEEtran}

\bibliography{mybib}

\end{document}